\documentclass[10pt,twocolumn,letterpaper]{article}

\usepackage{ijcb}
\usepackage{times}
\usepackage{epsfig}
\usepackage{graphicx}
\usepackage{amsmath}
\usepackage{amssymb}

\usepackage[numbers,sort,compress]{natbib}
\usepackage[norule,symbol,perpage]{footmisc}

\usepackage{booktabs}
\usepackage{multirow}
\usepackage{multicol}

\usepackage[pagebackref=true,breaklinks=true,letterpaper=true,colorlinks,bookmarks=false]{hyperref}

\ijcbfinalcopy % *** Uncomment this line for the final submission

\ifijcbfinal\fi

\makeatletter
\def\ps@IEEEtitlepagestyle{
\def\@oddfoot{\mycopyrightnotice}
\def\@evenfoot{}
}
\def\mycopyrightnotice{
{\hfill \footnotesize 978-1-6654-3780-6/21/\$31.00 \copyright 2021 IEEE\hfill}
}
\makeatother

\begin{document}

%%%%%%%%% TITLE
\title{Simultaneous Face Hallucination and Translation for Thermal to Visible Face Verification using Axial-GAN}

\author{Rakhil Immidisetti$^1$
\quad\quad\quad
Shuowen Hu$^2$

\quad\quad\quad
Vishal M. Patel$^1$\\\\
$^1$Johns Hopkins University
\quad
$^2$DEVCOM Army Research Laboratory\\

{\tt\small rimmidi1@jhu.edu, shuowen.hu.civ@mail.mil, vpatel36@jhu.edu}
}

\maketitle
\thispagestyle{empty}

%%%%%%%%% ABSTRACT
\begin{abstract}
   Existing thermal-to-visible face verification approaches expect the thermal and visible face images to be of similar resolution.
   This is unlikely in real-world long-range surveillance systems since humans are distant from the cameras. To address this issue, we introduce the task of thermal-to-visible face verification  from  low-resolution  thermal  images. Furthermore, we  propose Axial-Generative Adversarial Network (Axial-GAN) to synthesize high-resolution visible images for matching. In the proposed approach 
   we augment the GAN framework with axial-attention layers which leverage the recent advances in transformers for modelling long-range dependencies. We demonstrate the effectiveness of the proposed method by evaluating on two different thermal-visible face datasets. When compared to related state-of-the-art works, our results show significant improvements in both image quality and face verification performance, and are also much more efficient.
\end{abstract}

\let\thefootnote\relax\footnotetext{\mycopyrightnotice}

%%%%%%%%% BODY TEXT
\section{Introduction}

% \textit{Problem: Intro to problem. Thermal. Why low-res thermal is imp. \textbf{Have to expand intro, existing point}.} \textit{Distribution difference} \textbf{Didn't mention about synthesis :(}
%\textit{\textbf{Pre-justify variable resolution vs low-resolution}}

\begin{figure}[t!]
	\centering
	\includegraphics[width=\linewidth]{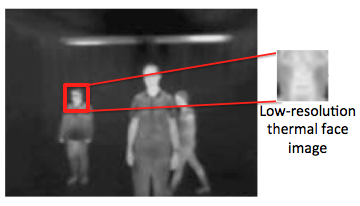}
	\caption{A typical thermal image \cite{Bilodeau_data}. Note that the captured images are of very low-resolution. In order to perform cross-modal face recognition, one needs to synthesize a high-resolution visible face image from a low-resolution thermal face image.}
	\label{fig:intro}
	\vspace{-5mm}
\end{figure}

In practical scenarios such as low-light or night-time conditions, one has to use thermal cameras for surveillance in order to detect and recognize faces. The acquired thermal images of faces in such scenarios have to be matched with existing biometric datasets that contain visible face images. Significant progress has been made by several works \cite{XingAttn,XingAttr,zhang2018synthesis,riggan2020,iranmanesh2018deep,ran1,ran2} to address the thermal-to-visible cross-spectrum face recognition problem. But existing works expect the thermal and visible face images to be of similar resolution. This is unlikely in real-world surveillance systems as humans are further away from cameras, thereby the region occupied by a face is much less when compared to an image in the visible face dataset. We illustrate the described issue in Figure \ref{fig:intro}. To address this, we introduce the task of matching low-resolution (LR) thermal face images against high-resolution (HR) visible face images.

%\textit{Additionally, thermal cameras are limited by their resolution. High-resolution thermal cameras have exponentially higher costs than low-resolution thermal cameras. \textit{Any citation??}
% }

% \textit{Brief summary of approach, motivation, intuition, advantages etc.  Description of our framework, aim: input output, gallery and probe verification}

\begin{figure*}[!tb]
	\centering
	\includegraphics[width=0.95\linewidth]{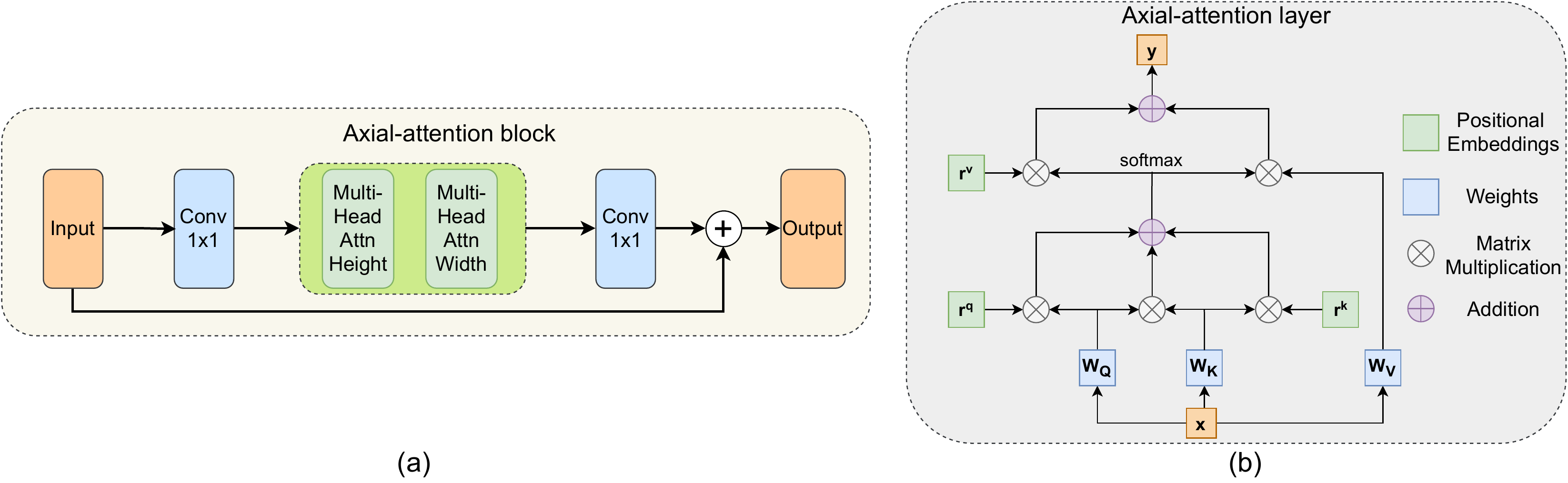}
	\caption{(a) The residual axial-attention block used in Axial-GAN.
	(b) Axial-attention layer, which is the basic building block of both height and width multi-head attention modules in the axial-attention block.}
	\label{fig:axial}
	\vspace{-5mm}
\end{figure*}

The large domain discrepancy between the thermal and visible images and the low resolution of the thermal images makes the introduced task quite challenging. To tackle it, we propose a hybrid network that augments an image-conditional generative adversarial network (GAN) \cite{GAN} with axial-attention \cite{wang2020axial}  layers.
The generator synthesizes face images in the visible domain, which are then matched against a gallery of visible images using an off-the-shelf face matching algorithm.
% Our developed hybrid network makes use of both convolutions and axial-attention layers to capture local and global information, respectively.
% , allowing for an efficient and effective attention-guided synthesis.
Using self-attention-based models \cite{transformer, SASA, wang2020axial} allows capturing the structural patterns of the face effectively, which is essential for tasks such as face verification.
However, stand-alone self-attention models require large-scale datasets for training.
Therefore, we develop a hybrid network that makes use of both convolutions and self-attention layers to efficiently capture the local and global information, respectively.
Additionally, augmenting our network with self-attention avoids the use of several stacked convolutional layers for modelling global dependencies. This makes our network extremely parameter efficient without any reduction in performance.
% Furthermore, we use spectral normalization and discriminator-based feature matching loss to stabilize the GAN training dynamics.
Although Di \etal \cite{XingAttn} proposed a similar hybrid network, it doesn't utilize positional information and multi-head design that are essential for capturing spatial structures and a mixture of features, which we incorporate into our network. To the best of our knowledge, this is one of the first works to propose a transformer-based GAN for face translation and face hallucination.

% \textit{(Better shift this to conclusion) The proposed network can also be used effectively for general image translation tasks. By just adding a couple of transformer based Axial blocks to existing architectures and using discriminator feature matching loss to stabilize the GAN training, existing baselines of image translation tasks can be improved.
% }

% \textit{Evaluation}

We evaluate our approach on the ARL-VTF dataset \cite{ArlVtf} and the polarimetric thermal face recognition dataset \cite{XingAttr}. We compare the performance of our approach with state-of-the-art methods in thermal-to-visible synthesis and also face hallucination. Our results show significant performance improvements in both image quality and face verification. Furthermore, an ablation study is conducted to demonstrate the effectiveness of axial-attention. Code is available at \url{https://github.com/sam575/axial-gan}.

%------------------------------------------------------------------------
\section{Related Work}

%\textbf{\textit{LR face recognition??}}

% As we are synthesizing a HR visible image from a LR thermal image, our problem involves both face hallucination and image translation. 
% To tackle the introduced problem, we propose an approach that draws on recent works in transformers and GANs.

%------------------------------------------------------------------------
\subsection{Thermal-to-visible face recognition}

%\textbf{\textit{Do I need to explain more about synthesis based GAN methods.}}

Several approaches have been proposed to address thermal-to-visible cross-spectrum face recognition, which can be mainly divided into two categories: feature-based and synthesis-based. Feature-based methods seek to find a common latent subspace where corresponding face images in each spectrum are closer in terms of some distance metric. Initial works used kernel prototype similarities \cite{klare2013heterogeneous}, partial least squares \cite{hu2015thermal,thermalfacerecognition2012} and coupled neural networks \cite{riggan2016optimal} on hand-crafted features 
such as  
% scale invariant feature transform (SIFT) or histogram of oriented gradient
SIFT and HOG.
Whereas, recent works leverage deep networks to extract domain-invariant features \cite{iranmanesh2018deep,riggan2020,ran1,ran2} or disentangled features \cite{disentangled2018}.

Synthesis-based methods have the advantage that they can leverage the recent advances in visible spectrum face recognition for matching the synthesized visible face images. Consequently, Riggan \etal \cite{Riggan2018thermal} used features from both global and local regions, and developed a region-specific cross-spectrum mapping for estimating visible images. Zhang \etal \cite{zhang2017generative,zhang2018synthesis} and Di \etal \cite{XingAttn,XingAttr} leverage GANs to enhance the perceptual quality of the synthesized images. 

%------------------------------------------------------------------------
\subsection{Face Hallucination}

Face hallucination is a domain-specific image super-resolution problem aimed at enhancing the resolution of a LR face image to generate the corresponding HR face image \cite{FhSurvey}. Consequently, most of the works exploit face-specific information such as attributes, landmarks, parsing maps, and identity for effective reconstruction of HR face images \cite{SuperFAN,SICNN,FSRNet}. This additional information is obtained either by human labelling or by using existing pre-trained models. Furthermore, to generate realistic faces many works tend to employ GANs
% as image super-resolution can be considered as an image-to-image translation task wherein LR images are translated into HR images, several works utilize GANs
\cite{FhGan,HiFaceGAN,FSRNet,SuperFAN}.

%------------------------------------------------------------------------
\subsection{Transformers}
% \begin{itemize}
% \item Attention allows to capture non-local information from entire feature maps which is conventionally achieved by stacking multiple convolutional layers.
% \item SAGAN is not efficient, no relative positional encodings, not a transformer, no mhsa.
% \end{itemize}
Transformers, introduced by \cite{transformer} leverage multi-head self-attention layers to compute pairwise correspondence between tokens to learn highly expressive features across long sequences. Recently, self-attention has been applied to many computer vision tasks such as classification, detection and segmentation  \cite{SASA,wang2020axial,valanarasu2021medical,NonLocal}. In contrast to non-local block models \cite{NonLocal,zhang2018self,XingAttn}, transformer-based models \cite{SASA,wang2020axial,valanarasu2021medical} use relative positional encodings and multi-head design, which is essential as they capture spatial structures in an image and a mixture of affinities, respectively.

%------------------------------------------------------------------------
\section{Proposed Method}

% \textit{Description of our framework, aim: input output, gallery and probe verification}

% \textit{Complete description of the overall method, network, losses, etc. Give overall structure.
% }

\begin{figure*}[!h]
	\centering
	\includegraphics[width=.99\linewidth]{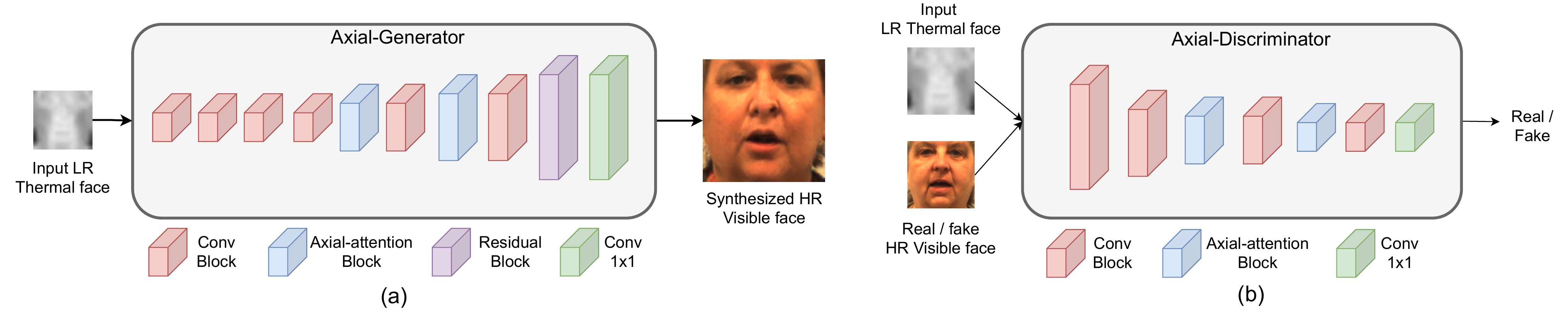}
	\caption{The proposed generator (a) and discriminator (b) augmented with axial-attention layers.}
	\label{fig:overall}
	\vspace{-5mm}
\end{figure*}

In this section, we discuss the details of the proposed Axial-GAN for thermal-to-visible synthesis from LR thermal images. In particular, we first give an overview of axial-attention \cite{wang2020axial, Ho2019AxialAI}, then discuss the proposed axial-attention-based generator and discriminator networks and finally address the objective functions and implementation details. 
% Furthermore, objective functions used for training and implementation details are also discussed.
% The overall framework is shown in Figure \c. 
% The generator, given a LR thermal image as input, synthesizes a HR visible image. In order to minimize the domain gap between the synthesized and real images, a patch-based GAN loss \cite{pix2pix2017} is used in addition to the perceptual loss \cite{johnson2016perceptual}. For stabilizing the training of GAN, we apply spectral normalization \cite{miyato2018spectral} to the networks and optimize the hinge version of adversarial loss \cite{lim2017geometric} and discriminator-based feature matching loss \cite{wang2018pix2pixHD}.

% \begin{figure*}[!htb]
% 	\centering
% 	\includegraphics[width=0.95\linewidth]{figures/overview.pdf}
% 	\caption{(a) The proposed axial-attention-based generator.
% 	(b) The residual axial-attention block used in Axial-GAN.
% 	(c) Axial-attention layer which is the basic building block of both height and width multi-head attention modules found in the axial-attention block.}
% 	\label{fig:overview}
% 	\vspace{-5mm}
% \end{figure*}

%------------------------------------------------------------------------
\subsection{Axial-attention overview}

% \textit{\textbf{Give brief of normal self-attention as well}. Say SA lacks MHSA blocks and positional encodings compared to Axial. Axial is also efficient allowing to use to on large feature maps. The Axial blocks are also residual.
%}

%\textbf{\textit{Multi-head, state equation?}}

Axial-attention \cite{Ho2019AxialAI,wang2020axial} factorizes 2D self-attention into two steps that apply 1D self-attention in height-axis and width-axis sequentially. Each step computes pairwise affinities, thereby learning a rich set of associative features across that particular axis. Combining them sequentially allows in capturing the full global information. The factorization helps in reducing the computational complexity and allows to capture long-range dependencies on larger regions, which is infeasible for 2D self-attention. Additionally, Wang \etal \cite{wang2020axial} incorporates relative positional bias \cite{Shaw2018SelfAttentionWR} in key, query and value as illustrated in Figure \ref{fig:axial} (b), thereby effectively associating information with positional awareness.
Formally, the position-sensitive axial-attention layer along the width-axis 
% as proposed by \cite{wang2020axial}
is as follows:
\begin{equation}\label{axial}
    y_{ij} = \sum_{w=1}^{W} \text{softmax}(q_{ij}^T k_{iw}+ q_{ij}^T r_{iw}^q + k_{iw}^T r_{iw}^k) (v_{iw} + r_{iw}^v),
\end{equation}
where $y_{ij}$ denotes the value at position $(i,j)$ in the output feature map. Here, $q, k, v$ denote query, key and value, respectively, which are computed from input feature map. $r^q, r^k, r^v$ denote the relative positional encoding for query, key and value, respectively. In our work, we adopt the axial-attention blocks as illustrated in Figure \ref{fig:axial} (a), which comprises of the multi-head position-sensitive axial-attention layers, to augment the GAN.
% for thermal-to-visible synthesis.

%------------------------------------------------------------------------
\subsection{Generator}

The generator, given a LR thermal image as input, synthesizes a HR visible image.
Inspired by the works in super-resolution \cite{wang2020sr-survey}, a progressive upsampling framework is used in the proposed generator as shown in Figure \ref{fig:overall} (a). This framework is efficient when compared to the pre-upsampling framework and also avoids the noise amplification caused by an upsampled input image.
% \textit{Inspired by the works in SR, a decoder type of generator is proposed in this work where we directly use the LR image instead of an image up-sampled using bicubic interpolation. This avoids adding any additional smoothing or artifacts caused by bicubic up-sampling} []. 
% The overall architecture of our generator is shown in Figure \ref{fig:overall} (a). 
Initially, we use only convolutional layers to learn local features such as edges, which are difficult to model by content-based mechanisms such as self-attention. In the later stages, we augment the network with axial-attention blocks to model the global context.
% Axial-attention blocks are added in the middle stages of the decoder. 
Additionally, in order to improve the stability of training, we use spectral normalization \cite{miyato2018spectral} for all convolutional layers except for those in the axial-attention block and the final output layer. Specifically, our generator architecture consists of the following components: 

\vspace{1mm}
\noindent
C64 - C128 - C256 - C512 - D256 - A256 - C256 - D128 - A128 - C128 - D64 - R64 - F
\vspace{1mm}

\noindent
where $Ck$, $Dk$, $Ak$ and $Rk$ denote $3\times3$ Convolution-BatchNorm-ReLU layer, $3\times3$ Deconvolution-BatchNorm-ReLU layer, axial-attention block and residual block \cite{he2016deep}, respectively, with $k$ filters.
$F$ denotes a $1\times1$ convolutional layer with Tanh as activation function, which produces a three channel output.

%------------------------------------------------------------------------
\subsection{Discriminator}

We use a patch-based discriminator \cite{pix2pix2017} augmented with axial-attention layers as shown in Figure \ref{fig:overall} (b), which is trained alternatively with the generator. The input to the discriminator is the concatenation of up-sampled thermal image and either real or fake visible image. Similar to the generator, spectral normalization is used for improving the training stability. 
% Axial-attention blocks are added in the middle stages of the discriminator. 
The discriminator consists of the following components:

\vspace{1mm}
\noindent
C64 - C128 - A128 - C256 - A256 - C512 - F
\vspace{1mm}

\noindent
where $Ck$ and $Ak$ denote a $4\times4$ Convolution-LeakyReLU layer and axial-attention block, respectively, with $k$ filters. $F$ denotes a $1\times1$ convolutional layer, which produces a single channel output.

%------------------------------------------------------------------------
\subsection{Objective function}

% \textit{Hinge gan (stability), perceptual loss, Importance of Discriminator feature loss}

The training dataset is given as a set of pairs $\{(x_i, y_i)\}$, where $x_i$ is a LR thermal image and $y_i$ is the corresponding HR visible image. We minimize the hinge version of adversarial loss \cite{lim2017geometric} for training the generator $G$ and discriminator $D$:
\begin{equation}\label{HingeLoss}
\begin{split}
% \begin{align}
    L_D =  &- \mathbb{E}_{x,y} [\min(0, -1 + D(x,y))]\\
    &- \mathbb{E}_{x} [\min(0, -1 - D(x,G(x)))]\\
    L_G =  &- \mathbb{E}_{x} [D(x,G(x))].
% \end{align}
\end{split}
\end{equation}
The overall loss function for the generator is defined as follows:
\begin{equation}\label{LossG}
L =  L_G + \lambda_H L_H + \lambda_P L_P + \lambda_{FM} L_{FM},
\end{equation}
\noindent
where $L_G$ is the adversarial loss for generator in Eq. \ref{HingeLoss}, $L_H$ is the Huber loss in Eq. \ref{LossH}, $L_P$ is the perceptual loss in Eq. \ref{LossP}, $L_{FM}$ is the discriminator-based feature matching loss in Eq. \ref{LossFM}. $\lambda_H,\ \lambda_P,\ \lambda_{FM}$ are the weights for Huber loss, perceptual loss and discriminator-based feature matching loss, respectively.

We use Huber loss between the target visible image and the synthesized visible image:
\begin{equation}\label{LossH}
L_H= 
\begin{cases}
    \mathbb{E}_{x,y} [0.5 * (G(x)-y)^2],& \text{if } |G(x) - y| < 1\\
    \mathbb{E}_{x,y} [|G(x)-y|] - 0.5,              & \text{otherwise}.
\end{cases}
\end{equation}

To generate visually pleasing results, perceptual loss \cite{johnson2016perceptual} is minimized, which is computed using features extracted from an off-the-shelf pre-trained VGG-19 network. The features from the initial layers help in generating high-frequency details, whereas the deeper layers help in enhancing the discriminative details.
The perceptual loss is formally stated below in which $F_i$ denotes the $i$-th layer of the VGG-19 network:  
\begin{equation}\label{LossP}
L_p = {E}_{x,y} [\|F_i(y) - F_i(G(x))\|_1].
\end{equation}

Additionally, discriminator-based feature matching loss \cite{wang2018pix2pixHD} is used to improve the training stability of GAN. Here, the $i$-th layer of discriminator $D$ is denoted as $D_i$:
\begin{equation}\label{LossFM}
L_{FM} = {E}_{x,y} [\|D_i(x,y) - D_i(x,G(x))\|_1].
\end{equation}

%------------------------------------------------------------------------
\subsection{Implementation}

% \textit{\textbf{Should I move this to experiments section?}}

% \textit{xavier initialization. Lr, Downsampling matlab, preprocessing, cropping, augmentation, hyperparameters. Image sizes. TTUR
% }

The entire network is trained on a single Nvidia 12 GB GPU. We choose $\lambda_H$ = 100 for the Huber loss, $\lambda_P$ = 10 for the perceptual loss and $\lambda_{FM}$ = 10 for the discriminator-based feature matching loss. Adam \cite{kingma2014adam} is used as the optimization algorithm with a learning rate of 0.0002 and the batch size is set to 32. The second convolutional layer at each scale in VGG-19 is used for the perceptual loss. The output features after each scale in the discriminator are used for the discriminator-based feature matching loss.

% \textit{128, random flip, Matlab, \textbf{Check Placement, Reasoning for down-sampling - lack of datasets?}}

The input to the generator is a $16\times16$ thermal image downsampled from the HR thermal image. The generator synthesizes the corresponding $128\times128$ visible face image. Downsampling or upsampling is performed using the MATLAB bicubic kernel function. Moreover, the training dataset is augmented with random horizontal flips.

%------------------------------------------------------------------------
\section{Experimental Results}

% \begin{table*}[!htb]
% \centering
% \begin{tabular}{lllllll}
% Input LR Thermal & pix2pix & SAGAN & HiFaceGAN & Ours & GT Visible & GT Thermal
% \end{tabular}
% \vspace{-5mm}
% \end{table*}
\begin{figure*}[!htb]
	\centering
	\includegraphics[width=0.95\linewidth]{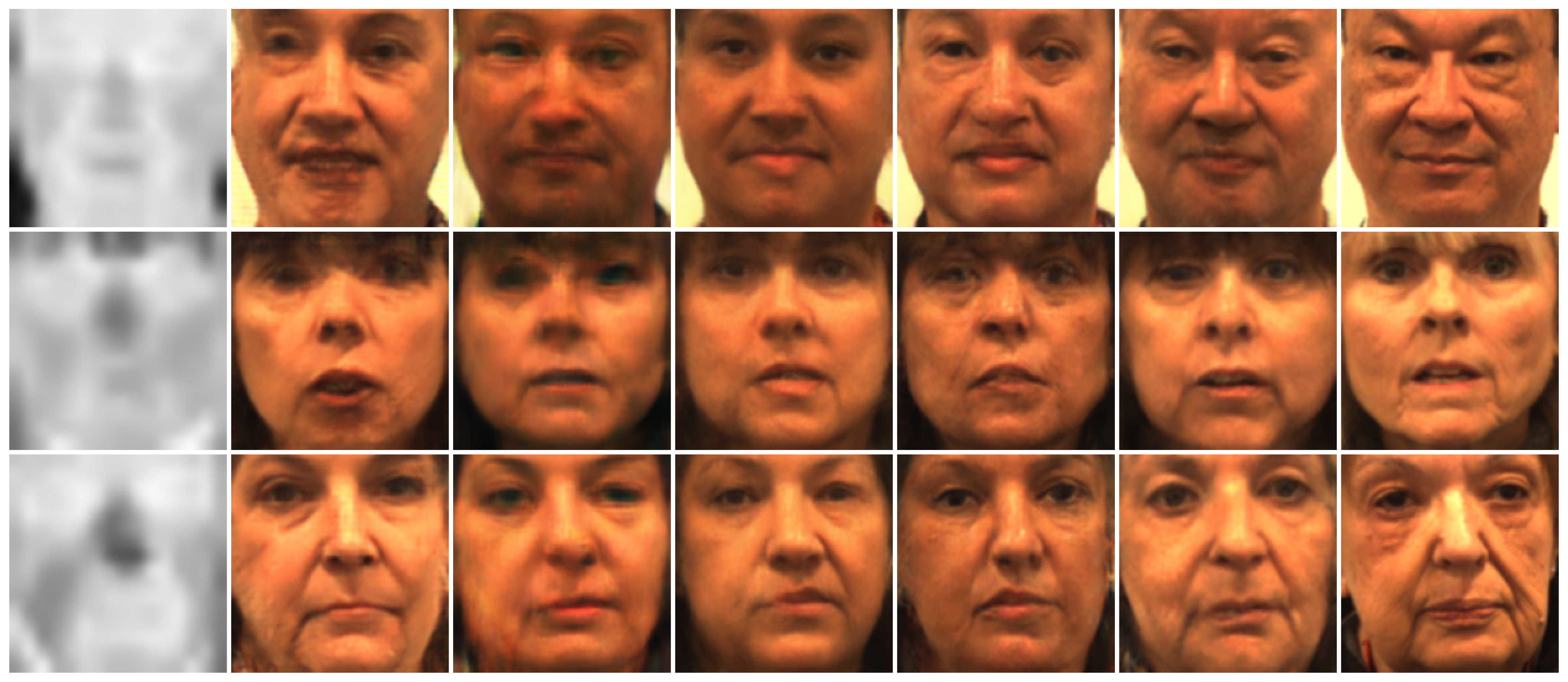}\\
	\raggedright
    \small
	\hspace{4.5mm} Input LR Thermal \hspace{4mm}
	pix2pix \cite{pix2pix2017} \hspace{6mm}
	SAGAN \cite{XingAttn} \hspace{3mm}
	HiFaceGAN \cite{HiFaceGAN} \hspace{8mm}
	Ours \hspace{8mm}
	Axial-GAN HR \hspace{2mm}
	GT HR Visible
	\vspace{1mm}
	\label{fig:vtf-qual}
	\caption{Synthesized visible images from different methods on the ARL-VTF dataset}
\end{figure*}

\begin{table*}[!htb]
\centering
\begin{tabular}{cccccccc}
\toprule
\multirow{2}{*}{Gallery} & \multirow{2}{*}{Method}                 & \multicolumn{2}{c}{P\_TB0-}     & \multicolumn{2}{c}{P\_TE0-}     & \multicolumn{2}{c}{P\_TB0+}     \\ \cmidrule(l){3-4}\cmidrule(l){5-6}\cmidrule(l){7-8} 
                         &                                         & AUC    (\%)        & EER (\%)           & AUC (\%)           & EER (\%)           & AUC    (\%)        & EER (\%)           \\ \midrule
\multirow{7}{*}{G\_VB0-} & pix2pix                                 & 91             & 17.77          & 88.94          & 19.37          & 80.06          & 28.13          \\
                         & SAGAN                                   & 92.29          & 15.3           & 90.78          & 17             & 79.18          & 28.38          \\
                         & HiFaceGAN                               & 91.29          & 17.13          & 89.42          & 18.49          & 82.82          & 26.21          \\
                         & Axial-GAN (Ours)                        & \textbf{94.4}  & \textbf{12.38} & \textbf{92.71} & \textbf{14.86} & \textbf{84.62} & \textbf{24.67} \\ \cmidrule(l){2-8} 
                         & Ours w\textbackslash{}o axial-attention & 90.89          & 17.75          & 88.87          & 19.64          & 80.07          & 27.69          \\
                         & Ours w self-attention                   & 92.72          & 14.62          & 90.52          & 17.29          & 82.15          & 26.44          \\
                         & Ours w HR Thermal                       & 99.05          & 4.98           & 98.07          & 7.1            & 91.45          & 17.85          \\ \midrule
\multirow{7}{*}{G\_VB0+} & pix2pix                                 & 86.32          & 22.43          & 83.7           & 24.58          & 91.15          & 17.89          \\
                         & SAGAN                                   & 87.8           & 22.27          & 85.85          & 23.93          & 87.37          & 21.44          \\
                         & HiFaceGAN                               & 86.12          & 23.7           & 84.24          & 25.01          & 91.03          & 16.61          \\
                         & Axial-GAN (Ours)                        & \textbf{89.71} & \textbf{19.75} & \textbf{88.01} & \textbf{21.58} & \textbf{93.62} & \textbf{14.05} \\ \cmidrule(l){2-8} 
                         & Ours w\textbackslash{}o axial-attention & 86.73          & 22.43          & 84.37          & 24.3           & 89.96          & 17.63          \\
                         & Ours w self-attention                   & 87.89          & 22.59          & 85.96          & 23.75          & 91.48          & 16.5           \\
                         & Ours w HR Thermal                       & 96.53          & 10.21          & 94.72          & 13.75          & 98.53          & 6.67 \\ \bottomrule    
\end{tabular}
\vspace{2mm}
\caption{Face verification results corresponding to the ARL-VTF Dataset}
\label{tab:vtf-verf}
\end{table*}

\begin{table*}[htp!]
	\centering
	\begin{minipage}{.4\textwidth}
		\centering
		\begin{tabular}{ccc}
        \toprule
        Method                         & PSNR            & SSIM          \\\midrule
        pix2pix                  & 16.243          & 0.549         \\
        SAGAN                    & 17.67           & 0.61          \\
        HiFaceGAN                & 17.764          & \textbf{0.62} \\
        Ours (Axial-GAN)         & \textbf{18.173} & 0.607         \\\midrule
        Ours w/o axial-attention & 17.067          & 0.577         \\
        Ours w self-attention    & 17.736          & 0.591         \\
        Ours w HR thermal        & 18.267          & 0.643        \\\bottomrule
        \end{tabular}
        \vspace{2mm}
        \caption{Image quality results on ARL-VTF dataset}
        \label{tab:vtf-psnr}
	\end{minipage}
	\begin{minipage}{.59\textwidth}
		\centering
		\begin{tabular}{cccccc}
        \toprule
        \multicolumn{1}{l}{Resolution}    & Method           & AUC           & EER           & PSNR           & SSIM          \\ \midrule
        \multirow{4}{*}{$24\times24$} & pix2pix          & 90.66          & 17.66          & 16.75          & 0.55          \\
                                 & SAGAN            & 92.26          & 15.72          & 17.66          & 0.60           \\
                                 & HiFaceGAN        & 91.32          & 17.09          & 18.21          & \textbf{0.63} \\
                                 & Ours (Axial-GAN) & \textbf{95.52} & \textbf{11.44} & \textbf{18.56} & \textbf{0.63} \\ \midrule
        \multirow{4}{*}{$8\times8$} & pix2pix          & 75.55          & 31.81          & 15.85          & 0.54          \\
                                 & SAGAN            & 72.62          & 33.81          & 16.30         & 0.53          \\
                                 & HiFaceGAN        & 75.71          & 31.61          & 16.36          & \textbf{0.57} \\
                                 & Ours (Axial-GAN) & \textbf{78.79} & \textbf{28.39} & \textbf{16.57} & 0.55          \\ \bottomrule
        \end{tabular}
        \vspace{2mm}
        \caption{Comparison of results for different resolutions on ARL-VTF dataset}
        \label{tab:res-ablation}
	\end{minipage}
\end{table*}

\begin{figure*}[t]
	\centering
	\begin{minipage}{.45\textwidth}
		\centering
		\includegraphics[width=.95\linewidth]{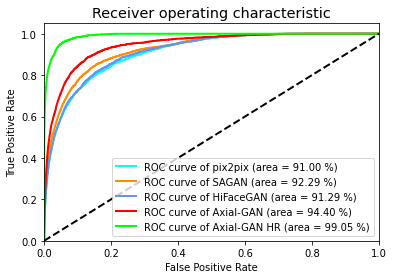}
    	\caption{The ROC curve comparison on ARL-VTF dataset}
    	\label{fig:vtf-roc}
	\end{minipage}
	\begin{minipage}{.45\textwidth}
		\centering
		\includegraphics[width=.95\linewidth]{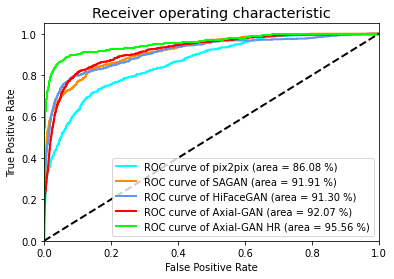}
	    \caption{The ROC curve for polarimetric thermal dataset}
	    \label{fig:arl3-roc}
	\end{minipage}
	\vspace{-2mm}
\end{figure*}

\begin{figure*}[]
	\centering
	\includegraphics[width=0.95\linewidth]{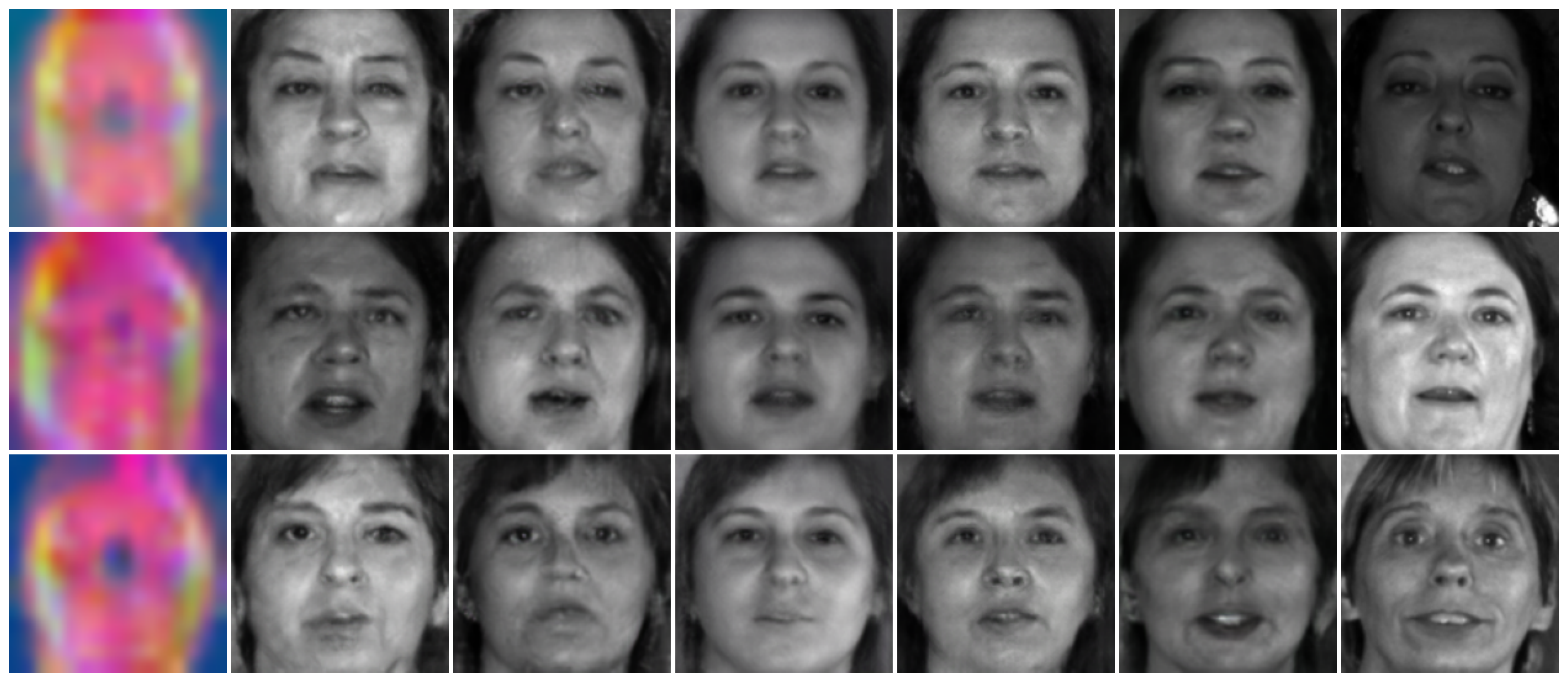}\\
	\raggedright
    \small
	\hspace{4.5mm} Input LR Thermal \hspace{4mm}
	pix2pix \cite{pix2pix2017} \hspace{6mm}
	SAGAN \cite{XingAttn} \hspace{3mm}
	HiFaceGAN \cite{HiFaceGAN} \hspace{8mm}
	Ours \hspace{8mm}
	Axial-GAN HR \hspace{2mm}
	GT HR Visible
	\vspace{1mm}
	\caption{Synthesized visible images from different methods on the ARL polarimetric thermal face dataset}
	\label{fig:arl3-qual}
\end{figure*}

\begin{table*}[]
\centering
\begin{tabular}{cccccc}
\toprule
Method           & AUC (\%)              & EER (\%)               & PSNR              & SSIM        & Parameters     \\ \midrule
pix2pix          & 81.573          $\pm$ 3.298 & 25.93           $\pm$ 2.733 & 16.488          $\pm$ 0.197 & 0.513          $\pm$ 0.021 & 41.8 M\\
SAGAN            & 84.377          $\pm$ 6.393 & 23.557          $\pm$ 6.699 & 17.329          $\pm$ 0.28  & 0.565          $\pm$ 0.016 & 7.9 M\\
HiFaceGAN        & 84.067          $\pm$ 6.801 & 22.947          $\pm$ 6.538 & 17.456          $\pm$ 0.176 & \textbf{0.583} $\pm$ 0.009 & 79.9 M\\
Ours (Axial-GAN) & \textbf{85.557} $\pm$ 5.307 & \textbf{22.347} $\pm$ 5.7   & \textbf{17.739} $\pm$ 0.165 & 0.582          $\pm$ 0.012 & 4.1 M\\ \midrule
Ours w HR Thermal   & 91.227          $\pm$ 4.17  & 15.53           $\pm$ 5.176 & 17.58           $\pm$ 0.05  & 0.588          $\pm$ 0.008 & 6.1 M\\ \bottomrule
\end{tabular}
\vspace{2mm}
\caption{Results for face verification and image quality on extended ARL multi-modal face recognition dataset}
\label{tab:arl3-quant}
\end{table*}

% \textit{How is the performance evaluated, protocols.
% PSNR metrics and verification metrics
% ROC curves, \textbf{Mention synthesized images for verf}
% }

We evaluate the performance of our method using image quality and face verification metrics. The quality of the synthesized visible images is evaluated using the peak signal-to-noise ratio (PSNR) and structural similarity (SSIM) index \cite{ssim}. The face verification performance is evaluated using the area under the curve (AUC) of receiver operating characteristic (ROC) and equal error rate (EER). Face verification scores are computed using cosine similarity between features extracted from `maxp\textunderscore5\textunderscore3' layer of a pre-trained VGG-Face model \cite{Parkhi2015VggFace}. The image quality metrics are evaluated on all images in the test set, whereas the face verification performance is evaluated on the protocols as described in each of the dataset section.

%------------------------------------------------------------------------
\subsection{Datasets and Protocols}

% \textit{Datasets New arl dataset, French dataset, Protocols. Preprocessing.}
% extended ARL polarimetric thermal face dataset

We use the ARL-VTF dataset \cite{ArlVtf} and the polarimetric thermal face recognition dataset \cite{XingAttr} for evaluating our method. We use only the baseline and expression images and ignore the pose images from these datasets in our evaluation, as it is extremely challenging to synthesize images from off-pose low-resolution images.\\

\noindent
\textbf{ARL-VTF dataset.} This is the largest dataset of paired conventional thermal and visible images, containing over 500,000 images from 395 subjects. The cropped face images are obtained using the provided bounding box annotations. The development and test split consist of 295 and 100 subjects, respectively.
For evaluating our method we follow the provided protocols, which prescribe different combinations of the gallery (G\_VB0-, G\_VB0+) and probe sets (P\_TB0-, P\_TE0-, P\_TB0+).
% which can be described by the following grammar.
% \textbf{\textit{Check placement.}}
Here, “G" and “P" denote the gallery and probe sets, respectively. Visible and thermal spectrum data are represented as “V” and “T”, respectively. “B” and “E” denote the baseline and expression sequences, respectively. “0" represents the images of subjects who do not possess glasses, while “-" and “+" represent the images of subjects who have their glasses removed or worn, respectively. For example, G\_VB0- is the set of visible images in the gallery where no subjects are wearing glasses.\\

\noindent
\textbf{Extended ARL multi-modal polarimetric thermal face recognition dataset.} This dataset contains a total of 5419 polarimetric thermal and visible image pairs corresponding to 121 subjects. A polarimetric thermal image consists of three Stokes images as its three channels: S0, S1, and S2, where S0 represents the conventional intensity thermal image, S1 represents the horizontal and vertical polarization-state information and S2 represents the diagonal polarization-state information. We follow the pre-processing steps outlined in \cite{XingAttr} for obtaining the cropped face images.
% \textit{The polarimetric thermal images are used as input to our network.}
The subjects are randomly divided into train, validation, and test sets containing 71, 25, and 25 subjects respectively.
Reported results are evaluated over three random splits. 
For evaluating thermal-to-visible face verification, the gallery set is formed using a random baseline visible image from each subject in the test set. The remaining disjoint polarimetric thermal images form the probe set.
%------------------------------------------------------------------------
\subsection{Results and Comparisons}
% \textit{Comparison with state-of-the art methods from I2I and SR. Qualitative and quantitative comments.}

We compare the performance of our method with pix2pix \cite{pix2pix2017}, Di \etal \cite{XingAttn} and HiFaceGAN \cite{HiFaceGAN}. Di \etal \cite{XingAttn} uses a self-attention \cite{NonLocal} based CycleGAN \cite{CycleGAN2017} (SAGAN) for thermal-to-visible synthesis. We use our implementation for comparison and ignore the cyclic-consistency part \ie, we do not train the GAN for visible-to-thermal synthesis for a fair comparison. HiFaceGAN is a recent work in face hallucination that uses content-adaptive convolutions to extract features for semantic guidance \cite{park2019SPADE} during replenishment. The competing methods use bicubic upsampled thermal images as input. Additionally, we compare with Axial-GAN that runs on HR thermal images to show the limitations of thermal-to-visible synthesis.

Table \ref{tab:vtf-verf} shows the thermal-to-visible verification performance of different methods on the ARL-VTF dataset. Compared to the other state-of-the-art methods, our method performs better with higher AUC scores and lower EER scores across all protocols. We also show the ROC curve for G\_VB0- \vs P\_TB0- protocol in Figure \ref{fig:vtf-roc}. Additionally, Table \ref{tab:vtf-psnr} shows the image quality results, where our method outperforms the other methods based on PSNR but is comparable to HiFaceGAN based on SSIM. Similar results can also be observed in Figure \ref{fig:vtf-qual}, which shows the synthesized visible images for all methods.
SAGAN has many artifacts in the synthesized images but the identity information is well retained.
HiFaceGAN synthesizes images that are smoother but the identity information is lost in this process. Our method comparatively synthesizes more realistic faces with well-defined face contours while preserving identity information. Furthermore, we conduct an additional study which shows the quantitative comparisons for different resolutions of input thermal image (see Table \ref{tab:res-ablation}). Here, we report the average AUC and EER scores across all protocols. We obtain results that are consistent with the results for $16\times16$ resolution. %\textbf{\textit{SAGAN failure, report architectural changes?}}

For the extended ARL multi-modal polarimetric face recognition dataset, the quantitative results are shown in Table \ref{tab:arl3-quant} and the ROC curve for one of the splits is shown in Figure \ref{fig:arl3-roc}. When compared to the ARL-VTF dataset, the performance of all methods decrease for this dataset. This is mainly because of its dataset size, which is 100 times lesser than that of ARL-VTF. Additionally, there is a lot of variation in illumination which can be seen in the visible images of Figure \ref{fig:arl3-qual}. As a consequence, there is a significant impact on the PSNR metric as thermal images fail to capture such variations. 
Our method performs better than the competing methods but the improvements are less when compared to the ARL-VTF dataset. 
HiFaceGAN performs relatively well when compared to its performance on the ARL-VTF dataset.
This can potentially be attributed to the larger dataset size required by the attention-based models for observing reasonable performance boosts.
Table \ref{tab:arl3-quant} also shows the parameters in the generator for each of the methods. Our method has approximately $10\times$, $2\times$, and $20\times$ lesser parameters than pix2pix, SAGAN, and HiFaceGAN, respectively.
Furthermore, Figure \ref{fig:arl3-qual} shows the qualitative comparisons of the competing methods. The qualitative observations for this dataset are consistent with that of the ARL-VTF dataset.\\
% Similar to the results in ARL-VTF dataset, HiFaceGAN synthesizes smoother

%------------------------------------------------------------------------
% \subsection{Ablation Study}
% \textit{w/o Axial block, replace with self-attention block. Effectiveness of discriminator with d-feat loss vs without vs normal.
% }
% \textit{make separate sub-section if loss is added?}

\noindent
\textbf{Ablation study.} In Table \ref{tab:vtf-verf} and \ref{tab:vtf-psnr} we also show the effectiveness of axial-attention. When we remove the axial-attention blocks from our method, we observe a significant decrease in performance.
This shows the importance of capturing global information using self-attention-based models.
Additionally, we also replace the axial-attention blocks with self-attention layers \cite{NonLocal,XingAttn} in our method. As expected, this performs similarly to SAGAN and falls short when compared to our method. This shows that the positional information and multi-head design are essential in improving the performance of self-attention-based models.
%------------------------------------------------------------------------
\section{Conclusion}

% \textit{What was proposed. Pointing out the improved results. Current importance of this paper and for future research as well.}

% \textit{Future work: Handling pose images, More experimentation of architectures for exploiting both transformer and convolutions. Stabilizing GAN training in transformers. \textbf{Variable size LR images}}

We introduced the task of thermal-to-visible face verification from low-resolution thermal images to deal with lower resolution faces in surveillance systems. To address this task, we proposed Axial-GAN in which we augment the GAN framework with axial-attention layers. Axial-attention effectively captures long-range dependencies with high efficiency. Our quantitative and qualitative results on multiple thermal-visible face datasets show improvements when compared to previous related works.
In future work, we would like to investigate the challenging task of face verification using off-pose LR thermal faces.
% In addition, synthesizing visible faces from off-pose thermal faces is a challenging task that needs further investigation.

%Incorporating axial-attention for other synthesis tasks such as semantic image synthesis \cite{park2019SPADE} and medical image synthesis \cite{JoseSynthesis} are potential directions for future research.

\section*{Acknowledgement}
This work was supported by ARO grant W911NF-21-1-0135.

{\small
\bibliographystyle{ieee}
\bibliography{ijcb2021}

\begin{thebibliography}{10}\itemsep=-1pt

\bibitem{Bilodeau_data}
G.~A. Bilodeau, A.~T. P.~L. St-Charles, and D.~Riahi.
\newblock Thermal-visible registration of human silhouettes: a similarity
  measure performance evaluation.
\newblock {\em Infrared Physics \& Technology}, 64:79--86, May 2014.

\bibitem{SuperFAN}
A.~Bulat and G.~Tzimiropoulos.
\newblock Super-fan: Integrated facial landmark localization and
  super-resolution of real-world low resolution faces in arbitrary poses with
  gans.
\newblock In {\em IEEE Conference on Computer Vision and Pattern Recognition},
  2018.

\bibitem{FSRNet}
Y.~Chen, Y.~Tai, X.~Liu, C.~Shen, and J.~Yang.
\newblock Fsrnet: End-to-end learning face super-resolution with facial priors.
\newblock In {\em IEEE Conference on Computer Vision and Pattern Recognition},
  2018.

\bibitem{thermalfacerecognition2012}
J.~Choi, S.~Hu, S.~S. Young, and L.~S. Davis.
\newblock Thermal to visible face recognition.
\newblock In {\em Proc.SPIE}, pages 8371 -- 8371 -- 10, 2012.

\bibitem{XingAttn}
X.~{Di}, B.~S. {Riggan}, S.~{Hu}, N.~J. {Short}, and V.~M. {Patel}.
\newblock Polarimetric thermal to visible face verification via self-attention
  guided synthesis.
\newblock In {\em 2019 International Conference on Biometrics (ICB)}, pages
  1--8, 2019.

\bibitem{XingAttr}
X.~{Di}, B.~S. {Riggan}, S.~{Hu}, N.~J. {Short}, and V.~M. {Patel}.
\newblock Multi-scale thermal to visible face verification via attribute guided
  synthesis.
\newblock {\em IEEE Transactions on Biometrics, Behavior, and Identity
  Science}, 3(2):266--280, 2021.

\bibitem{riggan2020}
C.~N. Fondje, S.~Hu, N.~J. Short, and B.~S. Riggan.
\newblock Cross-domain identification for thermal-to-visible face recognition.
\newblock In {\em 2020 IEEE International Joint Conference on Biometrics
  (IJCB)}, pages 1--9, 2020.

\bibitem{GAN}
I.~Goodfellow, J.~Pouget-Abadie, M.~Mirza, B.~Xu, D.~Warde-Farley, S.~Ozair,
  A.~Courville, and Y.~Bengio.
\newblock Generative adversarial nets.
\newblock In {\em Advances in Neural Information Processing Systems},
  volume~27, 2014.

\bibitem{he2016deep}
K.~He, X.~Zhang, S.~Ren, and J.~Sun.
\newblock Deep residual learning for image recognition.
\newblock In {\em IEEE conference on computer vision and pattern recognition},
  pages 770--778, 2016.

\bibitem{ran2}
R.~He, X.~Wu, Z.~Sun, and T.~Tan.
\newblock Learning invariant deep representation for nir-vis face recognition.
\newblock In {\em Proceedings of the AAAI Conference on Artificial
  Intelligence}, 2017.

\bibitem{ran1}
R.~{He}, X.~{Wu}, Z.~{Sun}, and T.~{Tan}.
\newblock Wasserstein cnn: Learning invariant features for nir-vis face
  recognition.
\newblock {\em IEEE Transactions on Pattern Analysis and Machine Intelligence},
  41(7):1761--1773, 2019.

\bibitem{Ho2019AxialAI}
J.~Ho, N.~Kalchbrenner, D.~Weissenborn, and T.~Salimans.
\newblock Axial attention in multidimensional transformers.
\newblock {\em arXiv preprint arXiv:1912.12180}, 2019.

\bibitem{hu2015thermal}
S.~Hu, J.~Choi, A.~L. Chan, and W.~R. Schwartz.
\newblock Thermal-to-visible face recognition using partial least squares.
\newblock {\em JOSA A}, 32(3):431--442, 2015.

\bibitem{iranmanesh2018deep}
S.~M. Iranmanesh, A.~Dabouei, H.~Kazemi, and N.~M. Nasrabadi.
\newblock Deep cross polarimetric thermal-to-visible face recognition.
\newblock In {\em 2018 International Conference on Biometrics (ICB)}, pages
  166--173, Feb 2018.

\bibitem{pix2pix2017}
P.~Isola, J.-Y. Zhu, T.~Zhou, and A.~A. Efros.
\newblock Image-to-image translation with conditional adversarial networks.
\newblock {\em CVPR}, 2017.

\bibitem{FhSurvey}
J.~Jiang, C.~Wang, X.~Liu, and J.~Ma.
\newblock Deep learning-based face super-resolution: A survey.
\newblock {\em arXiv preprint arXiv:2101.03749}, 2021.

\bibitem{johnson2016perceptual}
J.~Johnson, A.~Alahi, and L.~Fei-Fei.
\newblock Perceptual losses for real-time style transfer and super-resolution.
\newblock In {\em European Conference on Computer Vision}, pages 694--711,
  2016.

\bibitem{kingma2014adam}
D.~P. Kingma and J.~Ba.
\newblock Adam: A method for stochastic optimization.
\newblock In {\em International Conference on Learning Representations (ICLR)},
  2014.

\bibitem{klare2013heterogeneous}
B.~F. Klare and A.~K. Jain.
\newblock Heterogeneous face recognition using kernel prototype similarities.
\newblock {\em IEEE transactions on pattern analysis and machine intelligence},
  35(6):1410--1422, 2013.

\bibitem{lim2017geometric}
J.~H. Lim and J.~C. Ye.
\newblock Geometric gan.
\newblock {\em arXiv preprint arXiv:1705.02894}, 2017.

\bibitem{miyato2018spectral}
T.~Miyato, T.~Kataoka, M.~Koyama, and Y.~Yoshida.
\newblock Spectral normalization for generative adversarial networks.
\newblock In {\em International Conference on Learning Representations}, 2018.

\bibitem{park2019SPADE}
T.~Park, M.-Y. Liu, T.-C. Wang, and J.-Y. Zhu.
\newblock Semantic image synthesis with spatially-adaptive normalization.
\newblock In {\em IEEE Conference on Computer Vision and Pattern Recognition},
  2019.

\bibitem{Parkhi2015VggFace}
O.~M. Parkhi, A.~Vedaldi, and A.~Zisserman.
\newblock Deep face recognition.
\newblock In {\em BMVC}, 2015.

\bibitem{ArlVtf}
D.~Poster, M.~Thielke, R.~Nguyen, S.~Rajaraman, X.~Di, C.~N. Fondje, V.~M.
  Patel, N.~J. Short, B.~S. Riggan, N.~M. Nasrabadi, and S.~Hu.
\newblock A large-scale, time-synchronized visible and thermal face dataset.
\newblock In {\em IEEE Winter Conference on Applications of Computer Vision
  (WACV)}, 2021.

\bibitem{SASA}
P.~Ramachandran, N.~Parmar, A.~Vaswani, I.~Bello, A.~Levskaya, and J.~Shlens.
\newblock Stand-alone self-attention in vision models.
\newblock In {\em Advances in Neural Information Processing Systems}, 2019.

\bibitem{riggan2016optimal}
B.~S. Riggan, N.~J. Short, and S.~Hu.
\newblock Optimal feature learning and discriminative framework for
  polarimetric thermal to visible face recognition.
\newblock In {\em IEEE Winter Conference on Applications of Computer Vision
  (WACV)}, 2016.

\bibitem{Riggan2018thermal}
B.~S. Riggan, N.~J. Short, and S.~Hu.
\newblock Thermal to visible synthesis of face images using multiple regions.
\newblock In {\em IEEE Winter Conference on Applications of Computer Vision},
  2018.

\bibitem{Shaw2018SelfAttentionWR}
P.~Shaw, J.~Uszkoreit, and A.~Vaswani.
\newblock Self-attention with relative position representations.
\newblock In {\em NAACL-HLT}, 2018.

\bibitem{valanarasu2021medical}
J.~M.~J. Valanarasu, P.~Oza, I.~Hacihaliloglu, and V.~M. Patel.
\newblock Medical transformer: Gated axial-attention for medical image
  segmentation.
\newblock {\em arXiv preprint arXiv:2102.10662}, 2021.

\bibitem{transformer}
A.~Vaswani, N.~Shazeer, N.~Parmar, J.~Uszkoreit, L.~Jones, A.~N. Gomez,
  L.~Kaiser, and I.~Polosukhin.
\newblock Attention is all you need.
\newblock NIPS'17, page 6000–6010, 2017.

\bibitem{wang2020axial}
H.~Wang, Y.~Zhu, B.~Green, H.~Adam, A.~Yuille, and L.-C. Chen.
\newblock Axial-deeplab: Stand-alone axial-attention for panoptic segmentation.
\newblock In {\em European Conference on Computer Vision (ECCV)}, 2020.

\bibitem{wang2018pix2pixHD}
T.-C. Wang, M.-Y. Liu, J.-Y. Zhu, A.~Tao, J.~Kautz, and B.~Catanzaro.
\newblock High-resolution image synthesis and semantic manipulation with
  conditional gans.
\newblock In {\em IEEE Conference on Computer Vision and Pattern Recognition},
  2018.

\bibitem{NonLocal}
X.~Wang, R.~Girshick, A.~Gupta, and K.~He.
\newblock Non-local neural networks.
\newblock In {\em IEEE conference on computer vision and pattern recognition},
  pages 7794--7803, 2018.

\bibitem{wang2020sr-survey}
Z.~Wang, J.~Chen, and S.~C. Hoi.
\newblock Deep learning for image super-resolution: A survey.
\newblock {\em IEEE transactions on pattern analysis and machine intelligence},
  2020.

\bibitem{disentangled2018}
X.~Wu, H.~Huang, V.~M. Patel, R.~He, and Z.~Sun.
\newblock Disentangled variational representation for heterogeneous face
  recognition.
\newblock In {\em Proceedings of the AAAI Conference on Artificial
  Intelligence}, volume~33, pages 9005--9012, 2019.

\bibitem{HiFaceGAN}
L.~Yang, S.~Wang, S.~Ma, W.~Gao, C.~Liu, P.~Wang, and P.~Ren.
\newblock Hifacegan: Face renovation via collaborative suppression and
  replenishment.
\newblock In {\em 28th ACM International Conference on Multimedia}, pages
  1551--1560, 2020.

\bibitem{FhGan}
X.~Yu, F.~Porikli, B.~Fernando, and R.~Hartley.
\newblock Hallucinating unaligned face images by multiscale transformative
  discriminative networks.
\newblock {\em International Journal of Computer Vision}, 128(2):500--526,
  2020.

\bibitem{zhang2018self}
H.~Zhang, I.~Goodfellow, D.~Metaxas, and A.~Odena.
\newblock Self-attention generative adversarial networks.
\newblock {\em arXiv preprint arXiv:1805.08318}, 2018.

\bibitem{zhang2017generative}
H.~{Zhang}, V.~M. {Patel}, B.~S. {Riggan}, and S.~{Hu}.
\newblock Generative adversarial network-based synthesis of visible faces from
  polarimetrie thermal faces.
\newblock In {\em 2017 IEEE International Joint Conference on Biometrics
  (IJCB)}, pages 100--107, 2017.

\bibitem{zhang2018synthesis}
H.~Zhang, B.~S. Riggan, S.~Hu, N.~J. Short, and V.~M. Patel.
\newblock Synthesis of high-quality visible faces from polarimetric thermal
  faces using generative adversarial networks.
\newblock {\em International Journal of Computer Vision}, 127(6):845--862,
  2019.

\bibitem{SICNN}
K.~Zhang, Z.~Zhang, C.-W. Cheng, W.~Hsu, Y.~Qiao, W.~Liu, and T.~Zhang.
\newblock Super-identity convolutional neural network for face hallucination.
\newblock In {\em ECCV}, 2018.

\bibitem{ssim}
{Zhou Wang} and A.~C. {Bovik}.
\newblock A universal image quality index.
\newblock {\em IEEE Signal Processing Letters}, 9(3):81--84, 2002.

\bibitem{CycleGAN2017}
J.-Y. Zhu, T.~Park, P.~Isola, and A.~A. Efros.
\newblock Unpaired image-to-image translation using cycle-consistent
  adversarial networkss.
\newblock In {\em IEEE International Conference on Computer Vision (ICCV)},
  2017.

\end{thebibliography}
}

\end{document}